\documentclass[10pt,twocolumn,letterpaper]{article}
\usepackage[pagenumbers]{conf}   

\usepackage{graphicx}
\usepackage{amsmath}
\usepackage{amssymb}
\usepackage{booktabs}
\usepackage{multirow}
\usepackage{comment}

\definecolor{linkblue}{rgb}{0.21,0.49,0.74}
\usepackage[pagebackref,breaklinks,colorlinks,allcolors=linkblue]{hyperref}

\begin{document}

\title{Learning Sign Language Recognition under Label Noise: A Study of Noise-Robust Losses for Isolated and Continuous Settings}

\newcommand{\aff}[1]{\textsuperscript{#1}}
\newcommand{\affmark}[1]{\textsuperscript{#1}}

\author{
    Akihisa Shitara\aff{1} \quad
    Yoichi Ochiai\aff{1,2,3} \quad
    \\[2pt]
    {\normalsize \affmark{1}R\&D Center for Digital Nature, University of Tsukuba}
    \\[1pt]
    {\normalsize \affmark{2}Faculty of Library, Information and Media Science, University of Tsukuba} 
    \\[1pt]
    {\normalsize \affmark{3} Pixie Dust Technologies, Inc.}
    \\[2pt]
    {\tt\small theta-akihisa (at) digitalnature.slis.tsukuba.ac.jp}
}

\maketitle

\begin{abstract}
 In sign language recognition, the isolated-recognition (ISLR) classification loss treats a single label as ground truth, as does a frame-level auxiliary classifier over pseudo-labels---a component we add to continuous-recognition (CSLR) methods, which lack one. 
 Stylistic variation blurs ISLR annotation, and the absence of temporal boundaries in CSLR forces pseudo-labels; both are noisy, and multi-tier recognition---glosses and non-manual markers as separate tiers---would likewise require frame-level boundaries. 
 We therefore apply symmetric and generalized cross entropy (SCE, GCE)---robust alternatives to cross entropy (CE) established in image classification---not to connectionist temporal classification (CTC) but to the preceding single-label classifier. 
 On ASL Citizen with injected symmetric noise across three backbones (three seeds for ST-GCN), robust losses cost at most $2.5$~pt when labels are clean and beat CE by 2.9--10.0~pt in all six conditions at noise rate 0.2, one of which only after $q$ was re-selected on the dev set. GCE yields larger gains, but its optimal $q$ does not generalize across backbones, whereas a single SCE setting transfers to all nine conditions; both vary 2--11 times more than CE across runs, so a favorable point estimate does not establish stability. 
 For CSLR (PHOENIX-2014), we report no gain and instead probe how the type and effective weight of the auxiliary loss affect word error rate (WER). 
 Our frame-level targets exhibit a systematic assignment bias beyond natural boundary ambiguity, so the CSLR study constitutes a diagnosis of a single configuration. 
 Adding a pseudo-label CE auxiliary at $\lambda_{\mathrm{aux}} = 25$ to VAC, CorrNet, and SlowFastSign raises WER above the no-auxiliary baseline, and GCE/SCE improve on CE by 1.7--3.2~pt (three of six conditions return below that baseline). 
 However, at a common $\lambda_{\mathrm{aux}}$, the three losses differ by more than an order of magnitude in effective gradient, and under a control matching the initial gradient, the gap shrinks to 0.4--0.9~pt, while lowering the CE weight alone already beats that baseline. Neither the degradation nor the improvement can therefore be separated from the effect of the weight. 
 We use only symmetric noise, and multi-seed evaluation covers only ST-GCN and the VAC isolated configuration.

\end{abstract}

\section{Introduction}
\label{sec:intro}

 Sign language is a primary means of communication in Deaf communities, and its automatic recognition bears directly on information accessibility. 
 Deep learning-based sign language recognition~\cite{patel2024survey} is divided into isolated sign language recognition (ISLR), which classifies a single clip as a word, and continuous sign language recognition (CSLR), which recognizes a gloss sequence from unsegmented video. 
 ISLR is trained with a classification objective that treats the class label as ground truth; CSLR is trained with the connectionist temporal classification (CTC)~\cite{ctc} sequence loss; and staged methods additionally supervise the visual feature extractor with a frame-level classifier on pseudo-labels. It is the latter that we study, and since the CSLR methods we build on lack such a classifier, we add it ourselves. 
 The loss function itself, however, has rarely been the object of design.

 Sign language data nonetheless suffers from a label-noise problem that is easy to overlook. 
 In ISLR, the appearance of a sign varies substantially with signer, region, and style, so within-class variation is large and annotation blurs. 
 This is less a matter of labels being \emph{wrong} than of an uncertain supervisory signal due to expressive diversity, but at training time it behaves like label noise. 
 In CSLR, the issue is more direct. 
 Continuous data carries no temporal boundary label for each sign, so training the feature extractor with a frame-level cross entropy (CE) classifier requires pseudo-labels that assume a temporal extent for every gloss---either derived from the alignment of a sequence model~\cite{cui2019,pu2019} or, as in this work, obtained by assigning the gloss sequence uniformly along time. 
 Because real gloss durations are not uniform and transition intervals exist, such supervision is structurally noisy.

 \noindent\textbf{Being able to handle temporal boundaries matters beyond pseudo-label quality.} 
 Sign language annotation is multi-tiered: besides manual glosses, it describes non-manual markers (NMM: eyebrow raises, head movement, eye gaze, mouthing) on separate tiers, which may span several glosses when marking a question or a conditional clause, so handling multi-tier structure computationally presupposes a frame-level correspondence per tier~\cite{koller2020,camgoz2020}. 
 CTC avoids committing to boundaries by marginalizing over all admissible alignments and therefore does not say which frame belongs to which gloss. 
 A frame-level hard classification target is one of the few forms of supervision that can address this directly, yet, when used naively, it is fragile to noise; we study its robustification as a first step toward raising it to a level that multi-tier recognition can rely on. 
 (We evaluate word error rate (WER) only; boundary accuracy is not measured.)

 CE is known to be theoretically fragile under label noise~\cite{ghosh2017}, and image classification has produced many robust losses. 
 Generalized cross entropy (GCE)~\cite{gce} is defined as an interpolation between CE and the mean absolute error (MAE), and symmetric cross entropy (SCE)~\cite{sce} is a combination of CE with the reverse cross entropy (RCE); both are introduced in practice by replacing CE. 
 To the best of our knowledge, however, no systematic application of these losses to sign language recognition exists. 
 We aim to fill this gap and contribute the following.

 \begin{enumerate}
    \itemsep2pt
    \item We target label noise in sign language recognition---annotation blur from expressive diversity in ISLR, and boundary noise of pseudo-labels in CSLR---and evaluate SCE and GCE. 
    For ISLR, we inject symmetric noise at a known rate into ASL Citizen~\cite{aslcitizen} and test three skeleton- and appearance-based backbones under controlled conditions. 
    (Asymmetric noise, which would model a bias toward visually similar signs, is out of scope.)
    
    \item For CSLR, we use the pseudo-label CE classifier that supervises the feature extractor, not CTC, and analyze how the type and weight of the auxiliary classification loss affect WER. 
    At a fixed $\lambda_{\mathrm{aux}} = 25$, the CE auxiliary raises WER above the no-auxiliary baseline, and replacing it with SCE/GCE improves on this, in three backbones of differing strength (VAC, CorrNet, SlowFastSign).
    This degradation cannot, however, be attributed to the pseudo-labels themselves: under a gradient-matched control, lowering the CE weight without changing the loss shape already beats the no-auxiliary baseline (Sec.~\ref{sec:cslr}). 
    Nor is this a claim that we improve existing CSLR methods---with the original visual-enhancement/visual-alignment (VE/VA) configuration, the differences between losses nearly vanish, so our findings are confined to the choice of loss in configurations that use an auxiliary classification loss.

    \item \textbf{We give practical guidance for choosing a loss.} 
    Under noise, robust losses beat CE, but GCE yields larger gains. In contrast, its optimal $q$ transfers neither across noise rates nor across backbones, and in large-vocabulary training from scratch, a large $q$ makes the initial gradient vanish and falls below CE~\cite{staats2025}. In contrast, SCE yields more modest gains with a single setting that works across all conditions. 
    Furthermore, in an aspect not addressed by prior work, robust losses in our 2{,}731-class setting show a run-to-run standard deviation 2--11 times that of CE, showing that a favorable point estimate does not by itself imply operational stability (Sec.~\ref{sec:islr}; Appendix~C).

\end{enumerate}
\section{Related Work}
\label{sec:related_work}

\subsection{ISLR, CSLR and architectures}
 ISLR classifies a short single-sign clip into a word category~\cite{patel2024survey}, and has been addressed both with appearance-based video classifiers (I3D~\cite{i3d}, Video Swin Transformer (Video Swin-T)~\cite{videoswin}, MViTv2~\cite{mvitv2}, TimeSformer~\cite{timesformer}, VideoMAE~\cite{videomae}) and with skeleton-based methods (ST-GCN~\cite{stgcn}, SPOTER~\cite{spoter}); among the stronger methods is NLA-SLR~\cite{nlaslr}, which introduces language-aware label smoothing. 
 CSLR estimates a gloss sequence from unsegmented video and typically consists of a visual feature extractor, a sequence module, and an alignment module; because temporal boundaries are unavailable, CTC~\cite{ctc} is widely used for alignment. 
 VAC~\cite{vac} is the canonical framework that strengthens the visual extractor with auxiliary losses, and recent strong methods such as CorrNet~\cite{corrnet}, SlowFastSign~\cite{slowfastsign}, and AdaptSign~\cite{adaptsign} build on this VAC-style training. 
 CSLR should be distinguished from sign language translation (SLT), which maps a sign video to a spoken-language sentence and must resolve differences in word order and grammar rather than only recognize a gloss sequence. Since its formulation within neural machine translation~\cite{phoenix2014t}, work has learned recognition and translation jointly~\cite{camgoz2020slt}, back-translated monolingual text to enlarge the scarce parallel data~\cite{csldaily}, and transferred progressively from general-domain corpora~\cite{chen2022slt}; the state of the art is surveyed separately~\cite{decoster2024slt}. SLT is outside our scope.

\subsection{Learning with noisy labels and robust losses}
 Ghosh \etal showed that a loss is robust to label noise when it satisfies a symmetry condition, which MAE meets and CE does not~\cite{ghosh2017}; MAE, however, pays for this with \emph{slow learning and a potentially lower final accuracy}.

 Addressing this ``robustness versus trainability'' trade-off has driven subsequent work. 
 GCE introduces a parameter $q$ that bridges the two, treated as a single function that reduces to CE as $q\to 0$ and to MAE as $ q\to 1$~\cite {gce}. In contrast, SCE adds an RCE term to CE, so that the combination addresses both under-learning and overfitting in the presence of noise~\cite{sce}. 
 Later work proposes normalized losses and Active Passive Loss~\cite{apl}, as well as an asymmetry condition weaker than symmetry~\cite{asymmetric}.

 \noindent\textbf{Failures in large, many-class settings have also been reported.} 
 Ma \etal address the tendency of robust losses to under-fit through normalization~\cite{apl}, and Staats \etal show that on the 1{,}000-class WebVision robust losses that worked on CIFAR-10/100 largely fail to learn~\cite{staats2025}. 
 They attribute this to the poor overlap between the network's output at initialization and the region where gradients do not vanish, and mitigate it with a logit bias determined solely by the number of classes. 
 ASL Citizen, used here, has 2{,}731 classes and therefore falls in the regime where this problem can arise.

 \noindent\textbf{We do not use these improved variants, and evaluate GCE and SCE in the form given by their original papers.} 
 Since no systematic report exists for sign language recognition, establishing the behavior of the standard form is a prerequisite for evaluating improved variants.

\subsection{Loss design and training strategies in sign language recognition}
 Alternatives to CE do exist in sign language recognition. 
 For ISLR, language-aware label smoothing using semantic similarity between glosses has been proposed~\cite{nlaslr}, and gloss label inconsistency across merged corpora~\cite{wlasl} has been treated with consistent labeling and curriculum learning~\cite{neidle2022,dafnis2022}. 
 For CSLR, staged and iterative optimization with pseudo-labels~\cite{cui2019,pu2019} and auxiliary losses that strengthen the visual module, as in VAC~\cite{vac} and its successors~\cite{corrnet,slowfastsign}, are widely adopted against insufficient training of the visual extractor and the spikiness of CTC.
 Explicit noise-robust losses have rarely been applied to these auxiliary classification stages, and this is the gap we address. 
 The label smoothing in NLA-SLR softens the label distribution as a regularization, with a different motivation from suppressing the gradient contribution of mislabeled samples; a direct comparison is left for future work.

\section{Method}
\label{sec:method}

\begin{figure*}[t]
    \centering
    \resizebox{\textwidth}{!}{\includegraphics{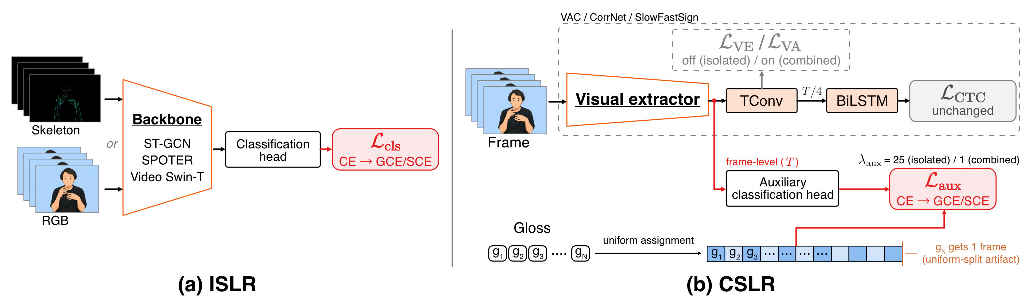}}
    \caption{Replacement points. 
    In (a) ISLR, we replace the loss at the classification head; in (b) CSLR, the auxiliary classification loss (\textcolor{red}{$\bigstar$}) on the visual-extractor output, before temporal downsampling (TConv). 
    \textbf{CTC is left unchanged in both settings.} 
    The three CSLR backbones differ only in the visual extractor, so the insertion point is identical across them. 
    VE/VA (dashed) are disabled in the isolated configuration and enabled in the combined one.}
    \label{fig:method}
\end{figure*}

\subsection{Loss definitions}
\label{sec:losses}
 For $K$-class classification, let $p(x)\in\mathbb{R}^{K}$ be the softmax output for input $x$, $y$ the one-hot target and $j$ the ground-truth class.

 \noindent\textbf{Cross entropy (CE):} $\mathcal{L}_{\mathrm{CE}} = -\sum_{k} y_k \log p_k(x)$.

 \noindent\textbf{Generalized cross entropy (GCE)}~\cite{gce}: for $q\in(0,1]$, $\mathcal{L}_{\mathrm{GCE}} = \bigl(1 - p_j(x)^{q}\bigr)/q$, which degenerates to CE as $q\to0$ and to MAE at $q=1$.

 \noindent\textbf{Symmetric cross entropy (SCE)}~\cite{sce}: with $\mathcal{L}_{\mathrm{RCE}} = -\sum_k p_k(x)\log y_k$, where $\log 0$ is clipped to a constant $A$ (we use $A = -4$, following the original paper and its public implementation), $\mathcal{L}_{\mathrm{SCE}} = \alpha\,\mathcal{L}_{\mathrm{CE}} + \beta\,\mathcal{L}_{\mathrm{RCE}}$.

 \noindent\textbf{Structural difference.} 
 GCE interpolates CE and MAE with a single function and replaces CE entirely, whereas SCE is an additive combination in which the CE term survives independently. 
 This difference governs the behavior when $q$ is large, or the coefficients are poorly set (Sec.~\ref{sec:islr}).

 All three are the original formulations defined over softmax probabilities $p_k$; the official ASL Citizen baseline~\cite{aslcitizen} instead uses a multi-label binary cross entropy, a setting in which we did not evaluate robust losses (limitation~(vii)).

\subsection{Application to ISLR}
 Because ISLR is a classification problem, we replace CE with GCE/SCE at the end of the classification head and leave the architecture unchanged. 
 The replacement applies using the same procedure to skeleton- and appearance-based backbones alike (Sec.~\ref{sec:backbones}), and we keep the formulation identical across architectures to prevent formulation differences from contaminating the comparison between backbones.

\subsection{Application to CSLR}
\label{sec:cslr-method}
 A central constraint of this work is that the CTC loss~\cite{ctc} is left unchanged. GCE and SCE are defined for single-label classification and cannot be substituted directly into the sequence marginalization of CTC. 
 Instead, our replacement target is the single-label CE classification loss over pseudo-labels, which supervises the feature extractor during staged/iterative training~\cite{cui2019,pu2019}. 
 The total loss is 
 
 \begin{equation}
    \mathcal{L} = \mathcal{L}_{\mathrm{CTC}} + \lambda_{\mathrm{aux}}\, \mathcal{L}_{\mathrm{aux}},
    \label{eq:total}
 \end{equation}

 and we swap the frame-level CE classification over pseudo-labels, $\mathcal{L}_{\mathrm{aux}}$, from CE to GCE/SCE.

 Pseudo-labels are generated by assigning the gloss sequence uniformly over time. 
 For a sequence of $T$ frames and $N$ glosses, frame $t$ receives gloss $\lfloor (N-1)\,t/(T-1) \rfloor$. 
 This rule distributes $N$ glosses over $N-1$ intervals, so the gloss boundaries drift late throughout the sequence, the drift accumulating toward the end (for $T=180$, $N=12$ it grows from $+2$ frames at the first internal boundary to $+14$ at the last, and 49\% of frames receive a gloss different from an exact equal split), and as its end point the final gloss is assigned only the last single frame. 
 Unlike generation from the CTC alignment of a sequence model~\cite{cui2019,pu2019}, this does not depend on the training state of the sequence model and provides a fixed target from the very beginning. 
 On the other hand, real gloss durations are not uniform, and transition intervals exist, so labels are systematically wrong near boundaries. This pseudo-supervision is therefore structurally noisy, which is the most natural justification for applying a robust loss. 
 We cite Cui \etal and Pu \etal only as sources of the pseudo-label CE method and do not reproduce their implementations (no publicly available implementations).

 We use VAC, CorrNet, and SlowFastSign as CSLR backbones. None of them has a plain CE classifier: VAC's visual enhancement (VE) is an auxiliary CTC ($\mathcal{L}_{\mathrm{VE}} = \mathcal{L}_{\mathrm{CTC}}^{\mathrm{v}}$) and its VA is a knowledge distillation using the Kullback--Leibler (KL) divergence, and CorrNet and SlowFastSign follow the same lineage with VE/VA-style auxiliary CTC and distillation. 
 We therefore add a pseudo-label CE auxiliary to each codebase, separately from VE/VA, and replace it with SCE/GCE. 
 We evaluate two configurations with different purposes.

 \noindent\textbf{(a) Isolated configuration.} 
 We turn off each method's own VE/VA and make the pseudo-label auxiliary the sole classification target of the visual encoder ($\lambda_{\mathrm{aux}} = 25$, reusing the weight $\lambda_{\mathrm{VA}} = 25$ that VAC applies to its VA term). 
 Removing the other visual supervision creates the condition under which differences between loss types are most visible; this is our primary object of analysis, and for VAC, we evaluate it with three seeds.

 \noindent\textbf{(b) Combined configuration.} 
 We keep each method's own VE/VA and add the auxiliary loss on top ($\mathcal{L}_{\mathrm{CTC}} + \mathcal{L}_{\mathrm{VE}} + \lambda_{\mathrm{VA}}\mathcal{L}_{\mathrm{VA}} + \lambda_{\mathrm{aux}}\mathcal{L}_{\mathrm{aux}}$, $\lambda_{\mathrm{aux}} = 1$), which checks whether the choice of loss affects final performance in a setting close to deployment.

 For both configurations, we also train a diagnostic condition without the auxiliary loss (isolated: $\mathcal{L}_{\mathrm{CTC}}$ only; combined: VE/VA only, i.e.\ the plain method), and use the latter to check our reproduction against published numbers. 
 CE serves as the baseline; label smoothing is not included in the comparison.

 Figure~\ref{fig:method} shows the replacement points in ISLR and CSLR. 
 In both settings, only the loss of a single-label classification head is replaced, and CTC is left untouched.

\subsection{Architectures}
\label{sec:backbones}
 To confirm that the effect of the replacement does not depend on a particular backbone, we use architectures of differing character in each task.

 For ISLR, we selected backbones in the following order. First, ST-GCN is used in the official ASL Citizen baseline, so comparing against the published numbers validates our implementation; the hyperparameter search is run on it.
 Next SPOTER, also skeleton-based but a Transformer rather than a graph convolution, which separates structural effects under an identical input representation. 
 Finally, the appearance-based Video Swin-T, which takes RGB frames, thereby confirming that the effect does not depend on a particular input modality. In all three, only the loss at the classification head is replaced.

 For CSLR, the starting point is VAC~\cite{vac}, the canonical framework on which later methods build, so the hyperparameter search is run on it; we then add CorrNet~\cite{corrnet}, and SlowFastSign~\cite{slowfastsign}, whose plain test WERs in our own training of the plain configuration are 21.1\%, 20.1\%, and 18.9\% (the published values differ slightly; see the supplementary material), so we can check whether the same tendency appears up to near-state-of-the-art performance. 
 Stronger ISLR architectures, skeleton-based CSLR, and replacing the sequence module are left for future work; I3D is excluded for the reason in limitation~(vii).
\section{Experiments}
\label{experiments}
 We do not treat all conditions uniformly in computational terms.
 The hyperparameter search is run only on ST-GCN for ISLR and on the VAC isolated configuration for CSLR; the selected values are transferred to the other architectures, and we check whether the effect is consistent.

\subsection{Setup}
\label{sec:setup}

 \noindent\textbf{Scope, metrics and splits.} 
 We address sign language recognition; translation is out of scope~\cite{decoster2024slt}, so our metrics are those of recognition, and we do not use BLEU. 
 ISLR uses Top-1/Top-5 accuracy (ASL Citizen is officially reported with Recall@$k$, which coincides with these since there is exactly one correct gloss); CSLR follows the official PHOENIX-2014~\cite{phoenix2014} protocol with NIST sclite including glm normalization. 
 We use the official splits, and ASL Citizen is evaluated only on signers unseen in training and validation.

 \noindent\textbf{Hyperparameter selection and the use of test.} 
 GCE requires $q$ and SCE requires $(\alpha, \beta)$. On the architectures we search, every setting is trained on the same train split and the best by the dev metric is selected---dev Top-1 for ISLR, dev WER for CSLR, the latter by official sclite with glm normalization. Epoch counts, early stopping and the search grid follow each method's official recipe (Appendix~A). \emph{Fallback rule}: when a transferred hyperparameter falls below CE on dev, we re-evaluate on dev the value selected at the other noise rates and adopt it. This fired exactly once, for GCE on SPOTER at noise 0.2, and is therefore a post hoc rescue of a single failing condition rather than a uniform protocol. Every decision is made on dev alone; test is used neither for hyperparameter selection, nor for epoch selection, nor for early stopping.

 \noindent\textbf{Controlled training conditions.} 
 To isolate the effect of the loss function, within a single architecture, the training settings, data augmentation, and initialization are held completely fixed across loss conditions; the optimizer and schedule follow each method's official configuration, and only the loss is changed.

 \noindent\textbf{Random seeds.} 
 ST-GCN for ISLR is trained on all nine conditions with three independent seeds (43/44/45), and we report the mean and standard deviation. 
 The training and noise-injection seeds are set to the same value, so initialization, data order and which labels are flipped all differ across replicates. The seed-42 search runs are excluded from the means because their dev score chose the setting. SPOTER, Video Swin-T and some CSLR conditions are single runs. 
 Determinism of GPU arithmetic is not enforced (\texttt{cudnn.benchmark=True}), so results vary between runs even at an identical seed and setting (up to 9.4~pt for SCE at noise 0.2 on ST-GCN). 
 The reported standard deviations are therefore run-to-run variance, containing both seed-induced variance and variation from the execution environment.

\subsection{ISLR on ASL Citizen}
\label{sec:islr}

 \noindent\textbf{Goal.} 
 To test, under controlled conditions, whether SCE/GCE improves on CE in the presence of label noise.

 \noindent\textbf{Dataset.} 
 ASL Citizen~\cite{aslcitizen} (American Sign Language; 2{,}731 signs, 83{,}399 videos, evaluated on unseen signers).

 \noindent\textbf{Noise setting.} 
 What we model is not adversarial mass corruption but naturally occurring label noise from annotation jitter and signer diversity, expected to be low in practice, so we restrict the noise rate to 0.0/0.1/0.2 (the 0.4--0.8 standard in the robust-loss literature is closer to an adversarial setting). 
 We inject symmetric noise at a known rate into the comparatively clean ASL Citizen, which isolates robustness as a function of the noise rate. 
 Asymmetric noise, biased toward visually similar signs, is closer to the generative process, but is out of scope here.

 \noindent\textbf{Architectures.} 
 We test CE/GCE/SCE against the noise rate on ST-GCN and SPOTER (skeleton), and on Video Swin-T (appearance); the search is run on ST-GCN and transferred to the other two. I3D is excluded for the reason in limitation~(vii).

 \noindent\textbf{Baseline.} CE is the reference. Label-smoothing CE is a regularization technique that softens the label distribution; its motivation differs from that of suppressing the gradient contribution of mislabeled samples, so it is not included here.

 \noindent\textbf{Results.} 
 Table~\ref{tab:islr} reports test Top-1 at the dev-selected hyperparameters.

\begin{table}[t]
    \centering\small
    \caption{ISLR test Top-1 (\%) under symmetric noise. ST-GCN, on which the search was run, uses three seeds (43/44/45; mean $\pm$ standard deviation, excluding the seed-42 sweep runs). SPOTER and Video Swin-T reuse the ST-GCN dev selection and are single runs; for GCE at noise 0.2 on SPOTER, $q=0.1$ by the fallback rule. Bold marks the best value per backbone and noise rate.}
    \label{tab:islr}
    \setlength{\tabcolsep}{4pt}
    \begin{tabular}{llccc}
        \toprule
            Backbone & Loss & noise 0.0 & noise 0.1 & noise 0.2 \\
        \midrule
            \multirow{3}{*}{ST-GCN}
            & CE  & \textbf{58.5}\,$\pm$\,0.3 & 46.3\,$\pm$\,0.9 & 36.2\,$\pm$\,0.4 \\
            & GCE & 56.0\,$\pm$\,1.2 & \textbf{51.8}\,$\pm$\,1.9 & \textbf{46.2}\,$\pm$\,4.3 \\
            & SCE & 57.0\,$\pm$\,1.4 & 50.6\,$\pm$\,3.4 & 44.8\,$\pm$\,3.7 \\
        \midrule
            \multirow{3}{*}{SPOTER}
            & CE  & 49.64 & 41.20 & 35.06 \\
            & GCE & \textbf{51.36} & \textbf{45.85} & \textbf{41.60} \\
            & SCE & 49.07 & 43.55 & 37.99 \\
        \midrule
            \multirow{3}{*}{Video Swin-T}
            & CE  & 58.42 & 53.27 & 46.42 \\
            & GCE & \textbf{59.14} & 54.66 & \textbf{55.06} \\
            & SCE & 57.91 & \textbf{55.88} & 53.32 \\
        \bottomrule
    \end{tabular}
    \par\vspace{2.5mm}
    \parbox{\linewidth}{\footnotesize GCE uses $q$=0.1 except ST-GCN / Video Swin-T at noise 0.2 ($q$=0.4); SCE uses $(\alpha,\beta)$=(0.1, 1.0) throughout.}
\end{table}

 \noindent\textbf{The cost when labels are clean is small}: the gap lies in $-2.5$ to $+1.7$~pt, and on ST-GCN it is comparable to the standard deviation of the same condition (1.2--1.4~pt), so a slight disadvantage under clean labels is visible but cannot be established at $n=3$.

 \noindent\textbf{Under noise, robust losses beat CE in every condition.} 
 They beat CE in all six conditions (three backbones $\times$ two losses) by $+1.4$ to $+5.5$~pt at noise 0.1 and by $+2.9$ to $+10.0$~pt at noise 0.2, the improvement being greater for GCE than for SCE on every backbone. One of the six---GCE on SPOTER at noise 0.2---attains this only after the fallback rule re-selected $q$ on dev; with the transferred $q=0.4$ it did not converge, so the six-of-six claim is conditional on the fallback rather than a property of GCE at a fixed hyperparameter. With $n = 3$ we run no significance test; what follows describes observed means.

 \noindent\textbf{The optimal $q$ does not transfer across backbones.} 
 On ST-GCN at noise 0.2, $q=0.1$ collapses to 29.41\% and a larger $q=0.4$ is required; on SPOTER, the transferred $q=0.4$ reached only 33.9\% on dev against 44.1\% for CE and did not converge, and $q=0.1$ is better. 
 The ST-GCN tendency that ``stronger noise favors a larger $q$'' does not hold across backbones.

 By contrast, the SCE hyperparameters transferred: $(\alpha, \beta) = (0.1, 1.0)$ worked across all three backbones at all three noise rates. 
 In practice, GCE yields larger gains if $q$ is chosen well, but incurs a per-backbone search cost, whereas SCE allows a single setting to be reused everywhere.

 \noindent\textbf{However, stability has two distinct senses, which do not coincide.} 
 In terms of robustness to changes in hyperparameters or backbones, SCE is the more stable option. 
 In terms of run-to-run variance at a fixed setting, however, the robust losses are clearly the least stable: the standard deviation of test Top-1 is 0.3--0.9~pt for CE against 1.2--4.3~pt for GCE and 1.4--3.7~pt for SCE (detailed in Appendix~C.2). 
 GCE at noise 0.2, in particular, spans 43.7--51.1\%, a range at which reporting a single run could lead to the wrong conclusion.

 \noindent\textbf{This does not match what the original papers report}, which give 0.05--0.39 (CE) and 0.01--0.42 (robust losses) at the same noise rate over 10--100 classes---a different level from our 3.7--4.3. 
 The cause cannot be attributed to the number of classes (dataset, architecture, and number of runs all differ); see Appendix~C.

 \noindent\textbf{The contrast between the two robust losses follows from their formulation.} 
 Because the CE term survives independently in SCE (Sec.~\ref{sec:losses}), trainability has a floor even when the RCE term contributes little, whereas GCE replaces CE entirely and has no term to fall back on once $q$ makes the gradient vanish---consistent with $q=0.4$ failing to converge on SPOTER while SCE worked. SCE is not unconditionally safe either: at $(\alpha,\beta) = (6.0, 1.0)$ it collapses at every noise rate, so the CE term acts as a floor only within an appropriate range of coefficients.

 \noindent\textbf{Some regions of the search space fail to train.} 
 Beyond the two ends of $q$ noted above, SCE collapsed outright (test Top-1 below 0.2\%) at $(\alpha, \beta) = (6.0, 1.0)$, so adopting a robust loss means the search range may contain regions in which learning does not take place. The best $q$ also varies with the noise rate, which cannot be estimated in advance for real data; the noisy-label literature addresses this by learning such hyperparameters via meta-learning~\cite{shu2020} or by estimating the noise rate~\cite{han2020,garg2025}.

 \noindent\textbf{Validity of the baseline.} Our clean ST-GCN Top-1 is $58.5 \pm 0.3$\%, 1.0~pt below the official ASL Citizen baseline of 59.52\% ($=$~R@1), likely because the official implementation trains in double precision. 
 Since our comparison fixes everything except the loss within a single architecture, this does not confound it. 
 We retain the dataset's preprocessing, which standardizes clips to 64 frames via length-dependent frame skipping rather than cropping, because some signs are compounds whose meaning can change with cropping.

\subsection{CSLR on PHOENIX-2014}
\label{sec:cslr}

 \noindent\textbf{Goal and dataset.} 
 We examine how the type and effective weight of the auxiliary classification loss over pseudo-labels affect WER on PHOENIX-2014~\cite{phoenix2014}, assessing generalization along the backbone axis rather than the dataset axis. 
 This is \emph{not} a comparison aimed at improving existing methods, and unlike ISLR, we inject no synthetic noise. Our pseudo-label generator also carries a systematic bias beyond natural boundary ambiguity (Sec.~\ref{sec:cslr-method}), so \textbf{everything in this section is a diagnosis of this particular configuration and does not generalize to pseudo-label supervision in CSLR at large.} 
 The pseudo-labels are obtained via uniform assignment (Sec.~\ref{sec:cslr-method}) and do not coincide with the true boundaries. The comparison here does not equalize the effective auxiliary pressure across losses, so the presence of noise cannot be inferred backward from the observed differences.

 \noindent\textbf{Configurations and hyperparameters.} 
 We evaluate both configurations of Sec.~\ref{sec:cslr-method} on three CSLR backbones of differing strength, together with a diagnostic line without the auxiliary loss (the no-auxiliary baseline) as the reference for its contribution. 
 Hyperparameters were searched for on the VAC isolated configuration ($q = 0.7$ for GCE, $(\alpha, \beta) = (0.1, 0.1)$ for SCE) and were transferred unchanged elsewhere. WER is computed by sclite with beam-search decoding.

\begin{table}[t]
    \centering\small
    \caption{CSLR, isolated configuration ($\mathcal{L}_{\mathrm{CTC}} + \lambda_{\mathrm{aux}}\mathcal{L}_{\mathrm{aux}}$, $\lambda_{\mathrm{aux}} = 25$). dev / test WER (lower is better).
    Bold marks the best test WER among the three losses, excluding the diagnostic. 
    VAC is the mean of three independent seeds (43/44/45) with the standard deviation on test; CorrNet and SlowFastSign are single runs.}
    \label{tab:cslr-isolated}
    \setlength{\tabcolsep}{4pt}
\begin{tabular}{lccc}
        \toprule
            Auxiliary loss & VAC (3 seeds) & CorrNet & SlowFastSign \\
        \midrule
            Diagnostic & 24.0 / 24.6 $\pm$ 0.8 & 21.1 / 21.6 & 19.9 / 20.2 \\
            CE & 25.9 / 26.5 $\pm$ 0.7 & 24.0 / 24.5 & 23.1 / 22.7 \\
            GCE & 22.9 / 23.9 $\pm$ 0.2 & 21.0 / \textbf{21.3} & 20.6 / \textbf{20.5} \\
            SCE & 22.6 / \textbf{23.5} $\pm$ 0.5 & 21.5 / 21.7 & 20.8 / 21.0 \\
        \bottomrule
    \end{tabular}
\end{table}

 \noindent\textbf{In all three backbones, GCE and SCE both beat CE}, by 1.7--3.2~pt in test WER---\textbf{but this margin is not attributable to the shape of the loss}, for the reason developed below. 
 On the three-seed VAC results, the seed-paired difference is $+2.60 \pm 0.92$~pt for CE\,$-$\, GCE and $+2.97 \pm 1.03$~pt for CE\,$-$\, SCE, with a consistent sign across all three seeds; with $n = 3$ we remain descriptive. 
 This cannot be taken as evidence that the pseudo-labels are noisy, since the three losses are compared at the same $\lambda_{\mathrm{aux}}$. At the same time, their effective auxiliary pressures differ by more than an order of magnitude (see below).

 \noindent\textbf{At the same time, the CE auxiliary is worse than using no auxiliary loss at all}, by 1.9--2.9~pt across the three backbones (against 0.7--0.8~pt of standard deviation for the two VAC conditions), even though auxiliary classification exists to offset the peakiness of CTC. 
 Robustification cancels this harm but does not greatly exceed the diagnostic line: the best robust loss is $-1.1$~pt (better) on VAC, $-0.3$~pt on CorrNet, and $+0.3$~pt (worse) on SlowFastSign, where the diagnostic line is in fact the best of the four conditions, so $\mathcal{L}_{\mathrm{CTC}}$ alone is already strong.

\begin{table}[t]
    \centering\small
    \caption{CSLR, combined configuration ($\mathcal{L}_{\mathrm{CTC}} + \mathcal{L}_{\mathrm{VE}} + \lambda_{\mathrm{VA}}\mathcal{L}_{\mathrm{VA}} + \lambda_{\mathrm{aux}}\mathcal{L}_{\mathrm{aux}}$, $\lambda_{\mathrm{aux}} = 1$). dev / test WER. 
    All single runs. Bold marks the best test WER among the three losses, excluding the diagnostic.}
    \label{tab:cslr-combined}
    \setlength{\tabcolsep}{4pt}
\begin{tabular}{lccc}
        \toprule
            Auxiliary loss & VAC & CorrNet & SlowFastSign \\
        \midrule
            Diagnostic & 20.8 / 21.1 & 19.5 / 20.1 & 18.2 / 18.9 \\
            CE & 20.6 / \textbf{21.4} & 19.4 / \textbf{20.0} & 18.1 / 18.7 \\
            GCE & 20.5 / 21.6 & 19.2 / \textbf{20.0} & 18.4 / 18.6 \\
            SCE & 20.4 / 21.8 & 19.5 / 20.6 & 17.9 / \textbf{18.4} \\
        \bottomrule
    \end{tabular}
\end{table}

 \noindent\textbf{With VE/VA present, the differences between loss types nearly vanish.} 
 In all three backbones, the four conditions, including the diagnostic line, fall within 0.7~pt (VAC 21.1--21.8, CorrNet 20.0--20.6, SlowFastSign 18.4--18.9). 
 The disadvantage of CE observed in the isolated configuration also disappears, and the ordering becomes unstable: CE is best on VAC, CE and GCE tie on CorrNet, and SCE is best on SlowFastSign---and on VAC and CorrNet, none of the three beats the diagnostic line.

 The training curves support this: $\mathcal{L}_{\mathrm{aux}}$ falls from 4.99 to 1.56 over 80 epochs in the isolated configuration but stalls at 5.16 to 3.65 in the combined one, so CTC and VE/VA dominate. 
 We therefore do not interpret differences between loss types in the combined configuration.

 \noindent\textbf{The WER improvement from GCE/SCE is not obtained by suppressing emissions.} 
 In the S/D/I breakdown, replacing CE with a robust loss reduces both substitutions and deletions, with insertions unchanged (see Appendix~B).

 \noindent\textbf{In CSLR, the tendency is reversed: the robust losses are the more stable.} 
 The run-to-run standard deviation over the four isolated conditions is 0.21 for GCE and 0.46 for SCE against 0.71 for CE and 0.81 for the diagnostic---well below the spread seen in ISLR (0.3--4.3~pt in Top-1), though the two are different metrics and not directly comparable. 
 The ordering SCE $<$ GCE $<$ diagnostic $<$ CE held for all three seeds (Appendix~C).

 \noindent\textbf{VE/VA contributes more than the auxiliary loss.} 
 Comparing the isolated with the combined diagnostic, WER improves by 3.5~pt on VAC, 1.5~pt on CorrNet, and 1.3~pt on SlowFastSign. Thus, VE/VA primarily determines CSLR performance, not the pseudo-label auxiliary loss; our contribution is limited to the choice of loss in configurations that use one. 
 The contrast is suggestive---VE and VA supervise the encoder without passing through a hard single label---but one cannot conclude that hard labeling is itself harmful: as shown next, the form of the target and the effective weight are confounded, and separating them would require a control that softens the labels at an identical effective weight, which we do not perform.

 \noindent\textbf{Only CE degrades monotonically as $\lambda_{\mathrm{aux}}$ grows, and it accelerates} ($+0.4$~pt $\rightarrow$ $+2.4$~pt over $1/10/25$ on the VAC combined configuration), whereas GCE and SCE are non-monotonic. 
 This cannot be read as ``robust losses tolerate weight settings better'', because the effective auxiliary pressure differs by more than an order of magnitude between losses at the same $\lambda_{\mathrm{aux}}$ (Appendix~C).

 \noindent\textbf{A gradient-matched control (VAC, isolated).} 
 The three losses share $\lambda_{\mathrm{aux}} = 25$, but they are unnormalized, and the gradient reaching the visual encoder differs in magnitude. At a uniform prediction ($K = 1{,}296$) the gradient on the ground-truth logit, $|\partial\mathcal{L}/\partial z_j|$, is 0.999 for CE, 0.0066 for GCE ($q=0.7$) and 0.100 for SCE ($\alpha=\beta=0.1$): at the same $\lambda_{\mathrm{aux}}$, CE receives roughly 150 times the auxiliary pressure of GCE and 10 times that of SCE. 
 We therefore added conditions in which $\lambda_{\mathrm{aux}}$ for CE is lowered so that the gradient at the start of training matches that of the robust losses. This control is run entirely with a single seed (42), and conditions are paired at equal effective weight $w = \lambda_{\mathrm{aux}} \cdot g$, where $g$ denotes the gradient magnitude given above (Table~\ref{tab:matched}).

\begin{table}[t]
    \centering\small
    \caption{Gradient-matched control (VAC, isolated), all with seed 42; test WER. Bold marks the better of each pair at equal effective weight $w$. The seed-42 values lie within the three-seed ranges of Table~\ref{tab:cslr-isolated}. The design is asymmetric: the GCE and SCE rows reuse the seed-42 sweep runs that selected those settings, whereas the CE rows are fresh; this favors the robust losses and, if anything, inflates the residual gap.}
    \label{tab:matched}
    \begin{tabular}{clcc}
        \toprule
            $w$ & Auxiliary loss & $\lambda_{\mathrm{aux}}$ & test WER \\
        \midrule
            0 & Diagnostic & 0 & 24.8 \\
            \textbf{0.17} & CE & 0.166 & 24.8 \\
            \textbf{0.17} & GCE & 25 & \textbf{23.9} \\
            \textbf{2.51} & CE & 2.508 & 23.4 \\
            \textbf{2.51} & SCE & 25 & \textbf{23.0} \\
            25.0 & CE & 25 & 26.7 \\
        \bottomrule
    \end{tabular}
\end{table}

 \noindent\textbf{The $\lambda_{\mathrm{aux}} = 25$ that VAC uses for VA is an excessive weight for CE}: in effective terms, it is $w = 25.0$, more than an order of magnitude from the other conditions, and it is at this point alone that CE falls below the diagnostic line (26.7 versus 24.8). 
 Lowering the weight removes the degradation without changing the loss shape---CE matches the diagnostic line at $w = 0.17$ (24.8) and beats it by 1.4~pt at $w = 2.51$ (23.4)---so an appropriately weighted pseudo-label CE auxiliary is not harmful, and is better than CTC alone.

 \noindent\textbf{Matching the effective weight shrinks the gap between losses considerably}, to 0.9~pt at $w = 0.17$ and 0.4~pt at $w = 2.51$. 
 Measured against CE at $\lambda_{\mathrm{aux}} = 25$, the apparent improvements were 2.8~pt (GCE) and 3.7~pt (SCE), so 1.9 and 3.3~pt of those were within reach by lowering the CE weight alone. 
 This is an arithmetic decomposition, not a separation into independent components: weight and loss shape interact through the gradient, and we matched only the gradient at initialization (as $p_j$ rises, the CE/GCE ratio shrinks from 150 to 1.6). 
 Our CSLR observation therefore remains at the level of ``with seed 42, lowering the CE weight alone attained the same WER range as the robust losses''; the residual 0.4--0.9~pt is comparable to the run-to-run standard deviation at $\lambda_{\mathrm{aux}} = 25$ (0.2--0.8~pt). The control is a single run per condition and was applied only to the VAC isolated configuration.

 \noindent\textbf{Reproduction.} Re-evaluating the official checkpoints through our pipeline reproduces the published dev/test WER exactly for VAC (21.2 / 22.3) and to within 0.1~pt for the other two, confirming the correctness of the evaluation pipeline; our own training of the combined diagnostic also comes close to the published values.

\section{Limitations}
\label{sec:limitations}

 \noindent(i) Multi-seed evaluation is confined to ST-GCN for ISLR (nine conditions) and the VAC isolated configuration for CSLR (four conditions), both at $n=3$; the other four backbones and the CSLR combined configuration are single runs, so the absolute size of the improvements may be biased.
 
 (ii) In the combined configuration, the auxiliary loss contributes little, and the non-monotonic per-loss behavior under varying $\lambda_{\mathrm{aux}}$ was measured only there, on VAC.
 
 (iii) Our CSLR implementation trains on a single GPU rather than distributed; without the auxiliary loss, all three backbones still reach a WER close to the published values.
 
 (iv) The hyperparameter search was run only on the VAC isolated configuration (CSLR) and on ST-GCN (ISLR), with the values transferred elsewhere, so some conditions miss their optimal settings.
 
 (v) The isolated configuration uses a common $\lambda_{\mathrm{aux}} = 25$ regardless of loss type, so the effective auxiliary pressure is not equalized; the gradient-matched control is a single run with seed 42 and matches only the gradient at initialization.
 
 (vi) The pseudo-label rule distributes $N$ glosses over $N-1$ intervals, so \textbf{the boundaries drift late throughout the sequence, the drift accumulating toward the end} (Sec.~\ref{sec:cslr-method}). 
 This may contribute to the auxiliary classifier being worse than no auxiliary at large $\lambda_{\mathrm{aux}}$, but \textbf{because the bias is common to all three losses, all three backbones, and both configurations, it does not explain the differences between loss types}. 
 The CSLR results are a diagnosis of this particular configuration and do not generalize to hard pseudo-label supervision in general.
 
 (vii) I3D was excluded from the main comparison because robust losses did not apply stably under the loss formulation of the official ASL Citizen baseline~\cite{aslcitizen} (a multi-label binary cross entropy).

 (viii) We report only average accuracy and do not measure per-signer disparity. Since the annotation blur we model originates in signer diversity, a loss that down-weights atypical samples could systematically under-serve signers whose style is least represented---a concern raised by Deaf-led critiques of research agendas in this area~\cite{desai2024}.

(ix) We do not compare against label-smoothing CE, excluded because softening the label distribution has a different motivation. That difference does not make the empirical comparison unnecessary, and label smoothing informed by inter-gloss similarity~\cite{nlaslr} would be a natural control for ISLR.

\noindent\textbf{On absolute performance.} 
 Following each architecture's official recipe, the clean Top-1 of Video Swin-T (58.42\%) falls below the official ASL Citizen I3D baseline (63.10\%): validation accuracy is still improving at the 30-epoch limit of that recipe. We evaluate the loss replacement within each architecture, not across backbones.
\section{Conclusion}
\label{sec:conclusion}

 We replaced the single-label classification target used in sign language recognition with noise-robust losses (SCE/GCE) across ISLR, CSLR, and several backbones. 
 Under injected noise on ASL Citizen, robust losses outperform CE across all six conditions at a noise rate of 0.2, with one exception: only after $q$ was re-selected on the dev set.  
 For CSLR, we treated the pseudo-label CE classifier added to VAC, CorrNet, and SlowFastSign as an object of diagnosis: at $\lambda_{\mathrm{aux}} = 25$, the CE auxiliary is worse than using none, and SCE/GCE improve on it, but \textbf{under a gradient-matched control lowering the CE weight alone attained the same WER range, so neither the degradation nor the improvement can be separated from the effect of the weight}, and our CSLR findings are confined to that configuration. 
 \textbf{Run-to-run variance behaves oppositely in the two tasks}---robust losses vary 2--11 times more than CE on ISLR but are more stable on CSLR---so neither observation should be generalized to the other task.

 Future work includes multi-seed evaluation with significance testing, asymmetric noise, multi-tier recognition with direct boundary evaluation, and mitigations for large-scale settings~\cite{staats2025,apl,asymmetric}, as well as extension to other benchmarks~\cite{phoenix2014t,csldaily} and to skeleton-based CSLR~\cite{cosign}.

\section*{Acknowledgements}
 This research was funded by Creatures Inc. 
 
 All AI-generated text and analysis outputs were read, verified, and revised by the authors, who take full responsibility for the content of this work.

{
    \small
    \bibliographystyle{ieeenat_fullname}
    \bibliography{refs}
}

\appendix
\section{Record of hyperparameter selection}
\label{app:sweeps}

 The body reports only the values selected on dev; every condition we explored is listed here. 
 The search was run on ST-GCN for ISLR and on VAC with the pseudo-label CE auxiliary for CSLR; the selected values were then transferred to the other architectures (Sec.~\ref{sec:setup}). 
 These sweeps show how the reported values were selected on dev; they are not a record of searching for a better setting afterwards. The seed-42 sweep runs are excluded from the means in the body to avoid the winner's curse.

\subsection{ISLR: sweep over $q$ for GCE}
 Values are test Top-1 (\%) / ECE, from the single seed-42 run used for selection; the main table in Sec.~\ref{sec:islr} retrains the selected value with three seeds and therefore differs. 
 Bold marks the condition selected on dev (validation Top-1). 
 We excluded $q=1.0$, for which vanishing gradients are known to prevent learning.

\begin{table}[htb]
    \centering\small
    \caption{ISLR, ST-GCN: sweep over $q$ for GCE. test Top-1 (\%) / ECE.}
    \setlength{\tabcolsep}{4pt}
    \begin{tabular}{lcccc}
        \toprule
            noise & CE & $q=0.1$ & $q=0.4$ & $q=0.7$ \\
        \midrule
            \multicolumn{5}{l}{\itshape test Top-1 (\%)}\\
            0.0 & 56.84 & \textbf{55.64} & 54.02 & 52.69 \\
            0.1 & 45.12 & \textbf{56.15} & 50.53 & 36.83 \\
            0.2 & 35.35 & 29.41 & \textbf{54.29} & 40.55 \\
        \midrule
            \multicolumn{5}{l}{\itshape ECE}\\
            0.0 & 0.259 & 0.292 & 0.318 & 0.353 \\
            0.1 & 0.327 & 0.260 & 0.349 & 0.479 \\
            0.2 & 0.127 & 0.435 & 0.321 & 0.441 \\
        \bottomrule
    \end{tabular}
\end{table}

\noindent\textbf{Performance degrades at both ends of $q$.} 
 At noise 0.1, $q=0.7$ falls below CE (36.83 against 45.12), which we attribute to vanishing gradients early in training (Appendix~\ref{app:analysis}). 
 Conversely, $q=0.1$ degrades to 29.41\% at noise 0.2: with a setting close to CE, overfitting to the noise dominates. 
 The optimum depends on the noise rate---0.1 at noise 0.0 and 0.1, 0.4 at noise 0.2.

\subsection{ISLR: sweep over $(\alpha,\beta)$ for SCE}
Same protocol as above. 
\begin{table}[htb]
    \centering\small
    \caption{ISLR, ST-GCN: sweep over $(\alpha,\beta)$ for SCE; test Top-1 (\%), single seed-42 runs. 
    Bold marks the condition selected on dev. 
    Values below 1\% indicate runs in which learning did not proceed.}
    \begin{tabular}{lccc}
        \toprule
            $(\alpha, \beta)$ & noise 0.0 & noise 0.1 & noise 0.2 \\
        \midrule
            CE (reference) & 56.84 & 45.12 & 35.35 \\
        \midrule
            $(0.1, 0.1)$ & 53.81 & 42.23 & 35.18 \\
            $(0.1, 1.0)$ & \textbf{57.42} & \textbf{50.41} & \textbf{40.30} \\
            $(6.0, 0.1)$ & 50.20 & 48.87 & 39.19 \\
            $(6.0, 1.0)$ & 0.11 & 0.09 & 0.18 \\
            $(10.0, 0.1)$ & 0.3 & 0.5 & 0.4 \\
            $(10.0, 1.0)$ & 0.6 & 0.7 & 0.9 \\
        \bottomrule
    \end{tabular}
\end{table}

\noindent\textbf{Raising $\alpha$ collapses learning except at $(6.0, 0.1)$.} 
 At $\alpha = 6.0$, every noise rate collapses when $\beta = 1.0$; with $\beta = 0.1$, the pairing the original paper used for CIFAR-100, learning does proceed---6.6~pt below CE on clean labels but 3.8~pt above it at both noise 0.1 and 0.2---though it stays short of the adopted $(0.1, 1.0)$ at every noise rate. 
 At $\alpha = 10.0$, all six conditions collapse. 
 Within the non-collapsing $\alpha = 0.1$ series, $(0.1, 1.0)$ is consistently best under noise.

 The $\alpha = 6.0$ series behaves peculiarly: test loss is very large (13--19), while ECE is very small for the non-collapsing $(6.0, 0.1)$ runs (0.03--0.04). The collapsed $(6.0, 1.0)$ runs are a separate case, with ECE near zero for the wrong reason (Appendix~\ref{app:analysis}). 
 The magnitude of the loss follows from the coefficient $\alpha = 6.0$ on the CE term. 
 That ECE is small, despite accuracy being well short of the adopted setting, suggests the output distribution is uniformly low-confidence, so that confidence and correctness diverge little; we did not verify this mechanism.

\noindent\textbf{Transfer.} 
 SPOTER and Video Swin-T were not searched; they reuse the dev-selected values above at each noise rate ($q=0.1$ at noise 0.0 and 0.1, $q=0.4$ at noise 0.2; $(\alpha,\beta) = (0.1, 1.0)$ throughout). 
 At noise 0.2 on SPOTER, the transferred $q=0.4$ fell below CE on dev, so the fallback rule of Sec.~\ref{sec:setup} replaced it with $q=0.1$.

\subsection{CSLR: sweep for the auxiliary loss}
 Run on the VAC isolated configuration only; the selected values were transferred to the other two backbones and to the combined configuration. sclite WER (\%).
 All entries below are single seed-42 runs, whereas the VAC column of Table~\ref{tab:cslr-isolated} is the mean of seeds 43/44/45; the dev values therefore differ slightly from those in that table.

\begin{table}[htb]
    \centering\small
    \caption{CSLR, VAC isolated: sensitivity to $q$ for GCE.}
    \begin{tabular}{lccc}
        \toprule
            $q$ & dev & test & test S / D / I \\
        \midrule
            0.1 & 25.4 & 25.7 & 14.4 / 8.1 / 3.1 \\
            0.4 & 23.4 & 24.0 & 13.1 / 8.1 / 2.7 \\
            \textbf{0.7} & \textbf{22.5} & \textbf{23.9} & 14.0 / 7.1 / 2.8 \\
            1.0 & 23.2 & 23.8 & 13.9 / 6.8 / 3.1 \\
        \bottomrule
    \end{tabular}
\end{table}

 Test WER is nearly flat at 23.8--24.0\% for $q = 0.4$--$1.0$; only $q = 0.1$ is clearly worse. 
 In the error breakdown, deletions decrease monotonically as $q$ increases ($D$: 8.1, 8.1, 7.1, 6.8), while substitutions remain roughly constant, so raising $q$ does not cause the model to withhold predictions. 
 Unlike ISLR, where both extremes of $q$ broke down, CSLR does not break down even at $q = 1.0$ (equivalent to MAE). 
 The CSLR auxiliary classifier operates on the gloss vocabulary (1{,}295 glosses; 1{,}296 classes, including the CTC blank), which is smaller than ISLR's 2{,}731 classes, and CTC provides a floor.

\begin{table}[htb]
    \centering\small
    \caption{CSLR, VAC isolated: all $(\alpha,\beta)$ for SCE.}
    \begin{tabular}{llccc}
        \toprule
            $\alpha$ & $\beta$ & dev & test & test S / D / I \\
        \midrule
            \textbf{0.1} & \textbf{0.1} & \textbf{21.6} & \textbf{23.0} & 13.3 / 6.9 / 2.9 \\
            0.1 & 1.0 & 23.8 & 24.8 & 13.7 / 8.1 / 3.0 \\
            6.0 & 0.1 & 30.4 & 31.0 & 17.8 / 9.4 / 3.8 \\
            6.0 & 1.0 & 30.0 & 30.5 & 16.5 / 11.4 / 2.7 \\
            10.0 & 0.1 & 31.4 & 31.3 & 16.7 / 11.6 / 3.0 \\
            10.0 & 1.0 & 30.8 & 30.6 & 16.2 / 11.9 / 2.5 \\
        \bottomrule
    \end{tabular}
\end{table}

 Relative to the adopted $(\alpha, \beta) = (0.1, 0.1)$, every high-$\alpha$ setting degrades WER by 7.5--8.3~pt, and both deletions ($D$: 6.9 to 9.4--11.9) and substitutions ($S$: 13.3 to 16.2--17.8) increase; holding $\beta$ fixed, raising $\alpha$ costs 8.0--8.3~pt at $\beta = 0.1$ and 5.7--5.8~pt at $\beta = 1.0$. The effect of $\beta$ is asymmetric across error types: lowering $\beta$ from 1.0 to 0.1 reduces deletions at every $\alpha$ ($\Delta D$ between $-0.3$ and $-2.0$). 
 At the adopted $\alpha = 0.1$, 1.2 of the 1.8~pt gained by lowering $\beta$ is accounted for by fewer deletions. 
 On ISLR (ST-GCN), by contrast, $\beta = 1.0$ was selected under noise, so the preferred direction of $\beta$ is opposite between the two tasks. 
 The error breakdown offers one explanation: the gain from lowering $\beta$ in CSLR comes mainly from fewer deletions, and deletions are penalized directly as a component of WER, whereas ISLR Top-1 has no analog of a deletion. 
 We did not test this hypothesis directly.

\noindent\textbf{On the validity of transferring.} 
 The $(\alpha, \beta) = (0.1, 0.1)$ fixed above was effective in the CorrNet isolated configuration (SCE 21.7\% against CE 24.5\%) but 0.3~pt worse than the $\beta = 1.0$ variant on SlowFastSign (test 20.7\%), and $\beta = 1.0$ was also better in the VAC combined configuration (21.0\% against 21.8\%). 
 We follow the pre-registered convention of transferring the value fixed on the isolated configuration, but that convention costs the best value in some conditions.

\subsection{Sensitivity to \texorpdfstring{$\lambda_{\mathrm{aux}}$} (VAC, combined)}
 Test WER when the weight of the auxiliary loss is varied--single runs.

\begin{table}[htb]
    \centering\small
    \caption{Sensitivity to $\lambda_{\mathrm{aux}}$, VAC combined configuration; test WER (\%). Bold marks the best value for each loss.}
    \begin{tabular}{lccc}
        \toprule
            Auxiliary loss & $\lambda_{\mathrm{aux}}{=}1$ & $\lambda_{\mathrm{aux}}{=}10$
            & $\lambda_{\mathrm{aux}}{=}25$ \\
        \midrule
            CE  & \textbf{21.4} & 21.8 & 23.8 \\
            GCE & 21.6 & 22.0 & \textbf{21.3} \\
            SCE & 21.8 & \textbf{21.4} & 21.8 \\
        \bottomrule
    \end{tabular}
\end{table}

 Only CE degrades monotonically ($+0.4$~pt then $+2.4$~pt relative to its own $\lambda_{\mathrm{aux}} = 1$ value); GCE and SCE are non-monotonic and fall below that reference at $\lambda_{\mathrm{aux}} = 25$ and $\lambda_{\mathrm{aux}} = 10$ respectively. 
 As noted in Sec.~\ref{sec:cslr}, this contrast cannot be read as robust losses tolerating the weight better, because the effective auxiliary pressure differs by more than an order of magnitude between losses at the same $\lambda_{\mathrm{aux}}$.

 The optimization pressure on the auxiliary loss also differs between the two configurations. 
 For the CE auxiliary on VAC, $\mathcal{L}_{\mathrm{aux}}$ falls from 4.99 at epoch 1 to 1.56 at epoch 80 in the isolated configuration ($\lambda_{\mathrm{aux}} = 25$), but only from 5.16 to 3.65 in the combined one ($\lambda_{\mathrm{aux}} = 1$), where CTC and VE/VA dominate.

\subsection{Reproduction check}
 We verified our implementation in two ways: by re-evaluating the officially released checkpoints through our evaluation pipeline, and by training the combined diagnostic (equivalent to the plain method) ourselves.

\begin{table}[htb]
    \centering\small
    \caption{Reproduction check; dev / test WER (\%). ``Ours'' is our own training of the combined diagnostic, which is equivalent to the plain method.}
    \setlength{\tabcolsep}{4pt}
    \begin{tabular}{lccc}
        \toprule
            Backbone & Reported & Official ckpt & Ours \\
        \midrule
            VAC & 21.2 / 22.3 & 21.2 / 22.3 & 20.8 / 21.1 \\
            CorrNet & 18.9 / 19.7 & 19.0 / 19.7 & 19.5 / 20.1 \\
            SlowFastSign & 18.0 / 18.3 & 18.1 / 18.4 & 18.2 / 18.9 \\
        \bottomrule
    \end{tabular}
\end{table}

 The official checkpoints reproduce the published values---exactly for VAC and within 0.1~pt for the other two---which confirms the correctness of the evaluation pipeline itself. 
 Our own training of the combined diagnostic is 0.4~pt behind on CorrNet and 0.6~pt behind on SlowFastSign, and 1.2~pt ahead on VAC; the plain test WERs quoted in Sec.~\ref{sec:backbones} (21.1, 20.1, 18.9) are these self-trained values, not the published ones.

\subsection{Reference values}
 The following are published numbers used only to check the strength of our baselines; they are not results of this work.

\begin{table}[htb]
    \centering\footnotesize
    \caption{Official baselines and reported values.}
    \setlength{\tabcolsep}{3.5pt}
    \begin{tabular}{llll}
        \toprule
            Task & Dataset & Method & Reported \\
        \midrule
            ISLR & ASL Citizen & I3D & R@1 63.10 / R@5 86.09 \\
            ISLR & ASL Citizen & ST-GCN & R@1 59.52 / R@5 82.68 \\
            CSLR & PHOENIX-2014 & VAC & dev 21.2 / test 22.3 \\
            CSLR & PHOENIX-2014 & CorrNet & dev 18.9 / test 19.7 \\
            CSLR & PHOENIX-2014 & SlowFastSign & dev 18.0 / test 18.3 \\
        \bottomrule
    \end{tabular}
\end{table}

\subsection{Schedules and search grids}
 The SCE grid was $\alpha \in \{0.1, 6.0, 10.0\} \times \beta \in \{0.1, 1.0\}$ in both tasks ($(0.1, 1.0)$ and $(6.0, 0.1)$ are the values the SCE paper used for CIFAR-10 and CIFAR-100; $\alpha = 10.0$ was added to probe outside that range), and the GCE grid was $q \in \{0.1, 0.4, 0.7\}$, with $1.0$ added for CSLR. 
 The ISLR search was run on ST-GCN (3 CE $+$ 9 GCE $+$ 18 SCE $=$ 30 conditions); Video Swin-T requires about 19 hours per condition, which is why it was not searched.

 Early stopping follows each method's official recipe. It therefore differs: ST-GCN and SPOTER run the full 100 epochs (\texttt{patience} disabled), Video Swin-T uses at most 30 epochs with patience 10, VAC and SlowFastSign at most 80 epochs with patience 40, and CorrNet at most 40 epochs with patience 40 (for CorrNet, the epoch count equals the patience, so early stopping never fires). 
 Every decision is made on dev alone; test is used for neither hyperparameter selection nor epoch selection, nor for early stopping.
 When synthetic noise is injected, dev is kept clean.
\section{Detailed analyses of the reported experiments}
\label{app:analysis}

 This section revisits calibration (ECE), the error breakdown (S/D/I), and the gradient, all computed from the same checkpoints and inference outputs as the experiments above.

\subsection{$q$ has both an upper and a lower bound}
 The GCE gradient is proportional to $-p^{q}(1-p)$, so a larger $q$ suppresses the influence of mislabeled samples but also makes the gradient vanish early in training. 
 With 2{,}731 classes, $p \approx 1/2731$ at initialization, and $p^{q}$ falls rapidly: $\approx 0.45$ at $q=0.1$, $\approx 0.04$ at $q=0.4$, $\approx 0.004$ at $q=0.7$. 
 This is why $q=0.7$ falls below CE on ST-GCN when labels are clean or nearly so, yet exceeds it at noise 0.2 (Appendix~\ref{app:sweeps}): whether a large $q$ pays off depends on the noise rate. 
 The failure of robust losses at large vocabularies is not specific to this work; the same tendency has been reported for the 1{,}000-class WebVision~\cite{staats2025}.

\subsection{Accuracy and calibration trade off}
 We computed the expected calibration error (ECE, 15 equal-width bins) for every condition in the sweep, and the setting selected by dev accuracy is not necessarily the best by ECE. 
 Averaged over three seeds at noise 0.0 / 0.1 / 0.2, ECE is 0.25 / 0.32 / 0.28 for CE, 0.28 / 0.31 / 0.37 for GCE, and 0.29 / 0.27 / 0.24 for SCE: as the noise rate rises, GCE becomes worse calibrated while SCE becomes better. 
 At noise 0.2, the most accurate model, GCE (46.2\%), is the worst calibrated (0.37), so accuracy and calibration are not jointly attainable. In applications such as dictionary retrieval, where a confidence is shown to the user, SCE may be a compromise between the two.

\begin{table}[htb]
    \centering\small
    \caption{ISLR, ST-GCN at noise 0.2: accuracy against calibration. 
    Single seed-42 runs, so that every setting in the sweep can be compared.}
    \begin{tabular}{lcc}
        \toprule
            Loss / setting & test Top-1 $\uparrow$ & ECE $\downarrow$ \\
        \midrule
            CE & 35.35 & 0.127 \\
            GCE $q=0.4$ (dev-best) & \textbf{54.29} & 0.321 \\
            GCE $q=0.7$ & 40.55 & 0.441 \\
            SCE $(0.1, 1.0)$ (dev-best) & 40.30 & 0.284 \\
            SCE $(6.0, 0.1)$ & 39.19 & \textbf{0.028} \\
        \bottomrule
    \end{tabular}
\end{table}

\noindent\textbf{A caveat.} 
 In settings where learning barely proceeds, the model never becomes confident, so ECE is small for the wrong reason: the collapsed $(6.0, 1.0)$ run has an ECE of 0.0001 while emitting an almost uniform distribution. 
 ECE alone must not be read as better calibration.

\subsection{The WER improvement is not obtained by withholding predictions}
 Because robust losses suppress the gradient contribution of low-confidence samples, a WER improvement could in principle arise from the model emitting fewer glosses---substitutions (S) falling while deletions (D) rise---reflecting caution rather than better discrimination. 
 We therefore inspected the sclite breakdown (isolated configuration, test, \%).

\begin{table}[htb]
    \centering\footnotesize
    \caption{Error breakdown, WER (S / D / I). VAC is the mean of three seeds; CorrNet and SlowFastSign are single runs. WER equals S$+$D$+$I up to rounding.}
    \setlength{\tabcolsep}{2.5pt}
    \begin{tabular}{lccc}
        \toprule
            Auxiliary loss & VAC (3 seeds) & CorrNet & SlowFastSign \\
        \midrule
            Diagnostic (CTC only) & 24.6 \tiny(14.17/7.67/2.77) & 21.6 \tiny(12.5/6.7/2.4)
             & 20.2 \tiny(11.8/6.3/2.2) \\
            CE & 26.5 \tiny(14.50/\textbf{9.13}/2.80) & 24.5 \tiny(13.1/9.0/2.4)
             & 22.7 \tiny(13.1/7.4/2.2) \\
            GCE \tiny($q$=0.7) & 23.9 \tiny(13.50/7.57/2.87) & 21.3 \tiny(12.6/6.5/2.2)
             & 20.5 \tiny(11.3/7.0/2.2) \\
            SCE \tiny($\alpha$=$\beta$=0.1) & 23.5 \tiny(13.57/\textbf{7.17}/2.70)
             & 21.7 \tiny(12.6/7.0/2.2) & 21.0 \tiny(11.9/7.1/1.9) \\
        \bottomrule
    \end{tabular}
\end{table}

\noindent\textbf{On VAC, the harm of the CE auxiliary appears mainly as more
deletions.} 
 Against the diagnostic line, deletions rise by $1.46$~pt while substitutions rise by only $0.33$~pt and insertions by $0.03$~pt.
 This composition differs by backbone, however: on CorrNet, deletions dominate ($+2.30$ against $+0.60$), whereas on SlowFastSign, substitutions are larger ($+1.30$ against $+1.10$). 
 A bias toward deletions cannot be stated as a general tendency.

\noindent\textbf{Replacing CE with a robust loss reduces deletions and substitutions together.} 
 For GCE, $\Delta S = -1.00$ and $\Delta D = -1.56$; for SCE, $\Delta S = -0.93$ and $\Delta D = -1.96$; insertions move by less than $0.1$~pt. 
 The pattern we were concerned about does not occur: both error types fall together.

\noindent\textbf{CE is also the most variable in deletions.} 
 The run-to-run standard deviation of deletions is 0.74~pt for CE, compared with 0.23 (diagnostic), 0.21 (GCE), and 0.15 (SCE): the CE auxiliary not only raises the level of deletion errors but also makes that level fluctuate between runs.

\noindent These are descriptive statistics at $n = 3$. 
 The three conditions share seeds in a paired design, but we did not aggregate per-seed differences across S/D/I, so we report per-condition means and standard deviations without significance tests. 
 Note also that when $\alpha$ was raised from 0.1 to 6.0/10.0 in the SCE sweep, deletions and substitutions both increased ($\Delta D$ between $+2.5$ and $+5.0$, $\Delta S$ between $+2.9$ and $+4.5$; Appendix~\ref{app:sweeps}), showing no such bias.

\subsection{Not attempted: synthetic noise for CSLR}
 Noise in CSLR is not confined to label flips. However, it includes structural ambiguity (boundary shifts, segment stretching, and substitution of similar glosses), which is costly to design separately from standard augmentation, so we evaluated it on the data as is. 
 The observed differences between losses, therefore, do not measure the amount of noise in the pseudo-label supervision; as noted in Sec.~\ref{sec:cslr}, they are equally explicable by differences in effective auxiliary pressure. 
 A control that varies the amount of noise is left for future work.

\subsection{Not attempted: side effects on fine-grained discrimination in ISLR}
 A model cannot distinguish a mislabeled sample from one that is correctly labeled but hard---such as a minimal pair differing only in handshape, movement, or location. 
 Robustification may therefore sacrifice discrimination precisely among the similar signs this work seeks to protect. 
 Similar pairs are a small fraction of the data, so an average metric like Top-1 would hide such degradation. 
 Verification by class-stratified accuracy or by a confusion rate restricted to similar pairs is left for future work.

\section{Extended discussion}
\label{app:discussion}

 This section elaborates on the interpretation of the observations above and contains no new experimental results.

\subsection{Points common to both settings}

\noindent\textbf{Annotation blur and the label quality of existing datasets.}
 Our ISLR motivation is that expressive diversity produces large within-class variation, so annotation blurs. 
 A separate problem is reported for existing ISLR datasets: gloss labeling is inconsistent across merged corpora---in WLASL, the same sign is labeled with different English glosses, and homographs are not distinguished~\cite{neidle2022,dafnis2022}. 
 The origins differ, but both make the ISLR supervisory signal uncertain. 
 We inject synthetic noise at a known rate into the comparatively clean ASL Citizen so that robustness can be studied as a function of the noise rate rather than of real noise of unknown magnitude.

\noindent\textbf{Diversity, representativeness and responsible research.} 
 The annotation blur we model ultimately originates in the diversity of signers, regions, and styles. 
 Deaf-led work has criticized sign language AI research for agendas set without the community~\cite{desai2024}. 
 Treating this diversity as ``noise'' to be suppressed risks discarding exactly the variation that makes the language what it is: a loss that down-weights atypical samples may systematically under-serve signers whose style is least represented. 
 Our evaluation reports only average accuracy and does not measure per-signer disparity, which we consider an important direction rather than a solved question

\noindent\textbf{Backbone choice and modality.} 
 ISLR covers skeleton-based (ST-GCN, SPOTER) and appearance-based (Video Swin-T) backbones, but all three CSLR backbones are appearance-based, so the CSLR conclusions are unverified for a skeleton-based visual extractor. 
 The set spans CNN, Transformer, and graph-convolutional structures but is not exhaustive: other GCN variants, large-scale self-supervised pre-training, and state-space models are untested. 
 The replacement applies as a swap of the classification head, so extending to them is straightforward in principle.

\noindent\textbf{Variation from the execution environment.} 
 We do not enforce deterministic GPU arithmetic (\texttt{cudnn.benchmark=True}), so identical seeds and settings still give different results between runs---up to 9.4~pt for SCE at noise 0.2 on ST-GCN. 
 The standard deviations we report therefore mix seed-induced with environment-induced variation; separating them would need deterministic execution or a nested design repeating each seed. 
 Multi-seed evaluation was affordable only for ST-GCN (about 40 minutes per condition) and the VAC isolated configuration; SPOTER (about 12 hours) and Video Swin-T (about 19 hours) were run once per condition.

\subsection{Points specific to ISLR}

\noindent\textbf{The noise rate is unknown in practice.} 
 We inject noise at a known rate, but a practitioner cannot estimate the noise rate of real data in advance, and the best $q$ moves with it. 
 The cost of guessing wrong is not small (Appendix~\ref{app:sweeps}), and the appropriate range depends on the backbone as well, so applying GCE to a new backbone should presuppose re-searching $q$, whereas SCE at $(\alpha, \beta) = (0.1, 1.0)$ worked in all nine conditions. 
 The noisy-label literature recognizes this: Shu \etal propose learning such hyperparameters via meta-learning~\cite{shu2020}, and estimating the noise rate itself has been established~\cite{han2020}, including dynamically during training~\cite{garg2025}.

\noindent\textbf{Restricted noise structure.} 
 We inject symmetric noise, the standard controlled condition in the robust-loss literature, whereas the process we posit---confusion toward visually similar signs---is closer to asymmetric noise. 
 The literature reports cases where the ordering of methods reverses between symmetric and asymmetric noise, so our ordering of GCE and SCE should not be assumed to carry over.

\noindent\textbf{Run-to-run variance.} 
 Standard deviations are 0.3--0.9~pt for CE against 1.2--4.3~pt for GCE and 1.4--3.7~pt for SCE, a ratio of 2.1--10.7 by condition; GCE at noise 0.2 spans 43.7--51.1\%. 
 The variance of the robust losses increases with the noise rate, whereas that of CE remains roughly constant. 
 This does not match the original papers, which report 0.05--0.39 (CE) and 0.01--0.42 (robust losses) at 10--100 classes, sometimes smaller than CE.

\textbf{The cause cannot be attributed to the number of classes.} 
 The GCE gradient coefficient at a uniform prediction, $p_j^{q} = K^{-q}$, is 0.042 in our noise-0.2 setting ($K = 2{,}731$, $q = 0.4$) and 0.040 in the CIFAR-100 setting of the original paper ($K = 100$, $q = 0.7$)---nearly identical initial gradients despite standard deviations differing by an order of magnitude (4.3 against 0.27). 
 Vanishing gradients alone, therefore, do not explain it, and comparisons across papers confound dataset, architecture, optimizer, and number of runs. 
 We did not run an ablation varying the class count alone, so what follows is a hypothesis. 
 In terms of \emph{accuracy}, a link between class count and the failure of robust losses is already reported~\cite{staats2025}; whether the variance we observe shares that mechanism is untested.

\textbf{The instability of SCE is not anticipated by its original paper.} 
 SCE is exceptionally stable there (0.04--0.19 at every noise rate) and its gradient is dominated by $\alpha(1-p_j)$, which does not vanish with the class count. 
 We used $(\alpha, \beta) = (0.1, 1.0)$, the setting the original paper applied to CIFAR-10; for CIFAR-100 it uses $(6.0, 0.1)$. 
 Since $\alpha = 0.1$ shrinks the classification gradient to a tenth of CE, the learning signal may be insufficient at 2{,}731 classes. 
 The CIFAR-100 pairing, which supplies more of it, does train but remains below the adopted setting at every noise rate, while raising $\alpha$ further or combining $\alpha = 6.0$ with $\beta = 1.0$ collapses learning outright. 
 The setting that transfers is therefore the CIFAR-10 one, and it is that setting's gradient scale we suspect of driving the variance; the appropriate range of $\alpha$ for a given class count remains open.

\noindent\textbf{Single-run reporting can mislead.} 
 The single-run values we initially had at noise 0.2 were CE 35.35\% and GCE 54.29\%, an apparent margin of $+18.9$~pt; the three-seed means give $+10.0$~pt. The GCE run in question is the very run whose dev score selected $q=0.4$ (the winner's curse), and it is excluded from the means. 
 The other two ISLR backbones and every CSLR condition outside the VAC isolated configuration are single runs and may exhibit the same bias.

\subsection{Points specific to CSLR}

\noindent\textbf{Weight balance and the scale gap between auxiliary losses.} 
 The weights in the VAC paper were tuned for a configuration using VE/VA only, and adding a pseudo-label CE auxiliary on top changes the balance among the terms.
 The isolated configuration shares $\lambda_{\mathrm{aux}} = 25$ across three unnormalized losses: at a uniform prediction, the gradient on the ground-truth logit is 0.999 for CE, 0.0066 for GCE, and 0.100 for SCE, so CE receives roughly 150 times the auxiliary pressure of GCE and 10 times that of SCE.

\noindent\textbf{The gradient-matched control has its own limits.} 
 We matched only the gradient at initialization; as $p_j$ rises, the CE/GCE ratio shrinks from 150 to about 1.6, so the match does not hold during training. 
 The control is a single run per condition at one seed, was run only on the VAC isolated configuration, and its residual gap of 0.4--0.9~pt is comparable to the run-to-run standard deviation measured at $\lambda_{\mathrm{aux}} = 25$ (0.2--0.8~pt). 
 A cleaner design would select the weight of each loss on dev, or match gradient norms across all backbones and several seeds throughout training.

\noindent\textbf{On verifying the mechanism.} 
 We show a correspondence between robustifying the auxiliary loss and improving WER; the mechanism---that the visual encoder learns a better representation---is inferred rather than demonstrated. 
 A direct test would train the visual encoder alone as a classifier, detached from the sequence module; we did not do this.

\noindent\textbf{Scope of the contribution.} 
 What primarily determines CSLR performance is VE/VA, not the pseudo-label auxiliary we study; the isolated and combined diagnostics differ by 3.5, 1.5, and 1.3~pt across the three backbones.
 WER is not the only axis, either. 
 VE and VA supervise at the sequence and distribution level and do not say which frame corresponds to which gloss, whereas moving toward multi-tier recognition requires supervision that can carry per-tier boundaries. 
 A frame-level hard classification target is one of the few candidates. However, this work says nothing about the quality of the boundaries it produces: our pseudo-labels are assumed to be uniform rather than estimated, and we measure neither boundary accuracy nor an NMM tier.

\noindent\textbf{Ways of obtaining boundaries differ.} 
 Our uniform assignment is distinctive in that its boundaries are fixed and never updated, unlike staged and iterative optimization, which periodically refreshes the alignment~\cite{cui2019,pu2019}, or multi-stream HMMs, which additionally handle parallel streams under explicit synchronization constraints~\cite{koller2020}.
 Handling a multi-tier structure in earnest requires that last axis. Refining boundary estimation, however, does not remove the noise. Those frameworks remain weakly supervised, and the assigned labels remain uncertain near boundaries, so robustification and better alignment are plausibly complementary rather than alternatives.

\noindent\textbf{Run-to-run variance in CSLR.} 
 Across the four isolated conditions, the standard deviation is 0.21--0.81~pt, well below the spread seen in ISLR (0.3--4.3~pt in Top-1)---though the two are different metrics and not directly comparable---and the robust losses are more stable (GCE 0.21, SCE 0.46, against CE 0.71 and diagnostic 0.81). 
 The ordering SCE $<$ GCE $<$ diagnostic $<$ CE held for all three seeds, and the seed-paired differences are $+2.60 \pm 0.92$~pt for CE\,$-$\, GCE and $+2.97 \pm 1.03$~pt for CE\,$-$\, SCE. 
 This is the opposite of the ISLR observation, and neither should be generalized to the other task.

\end{document}